\documentclass{bmvc2k}
\def\theirs{LAttQE\xspace}

\def\ours{GQE\xspace}
\def\oursfull{Graph Query Expansion\xspace}

\definecolor{greenn}{rgb}{0.30,0.69,0.31}
\definecolor{greennn}{rgb}{0.10,0.50,0.10}
\definecolor{yelloww}{rgb}{1.0000,0.8392,0}
\definecolor{reddd}{rgb}{0.70,0.20,0.20}
\definecolor{pinkkk}{rgb}{0.898,0.0,0.48218}

\def\roxf{$\mathcal{R}$Oxford\xspace}
\def\rox{$\mathcal{R}$Oxf\xspace}

\def\rpar{$\mathcal{R}$Paris\xspace}
\def\rpa{$\mathcal{R}$Par\xspace}

\def\rmil{$\mathcal{R}$1M\xspace}

\usepackage{times}
\usepackage{epsfig}
\usepackage{graphicx}
\usepackage{amsmath}
\usepackage{amssymb}
\usepackage{color, colortbl}
\usepackage{xcolor}
\usepackage{multirow}
\usepackage{booktabs}
\usepackage{multicol}%
\usepackage{floatrow}
\usepackage{subfig}
\usepackage[font=footnotesize,skip=0pt]{caption}
\usepackage[ruled,vlined,linesnumbered]{algorithm2e}
\usepackage{float}

\let\oldnl\nl
\newcommand{\nonl}{\renewcommand{\nl}{\let\nl\oldnl}}

\newlength\mylen
\newcommand\myinput[1]{%
  \nonl
  \settowidth\mylen{\KwIn{}}%
  \setlength\hangindent{\mylen}%
  \hspace*{\mylen}#1\\}

\mathchardef\mhyphen="2D
\mathchardef\mplus=\mathcode`+
\newcommand*{\belowrulesepcolor}[1]{%
  \noalign{%
    \kern-\belowrulesep
    \begingroup
      \color{#1}%
      \hrule height\belowrulesep
    \endgroup
  }%
}
\newcommand*{\aboverulesepcolor}[1]{%
  \noalign{%
    \begingroup
      \color{#1}%
      \hrule height\aboverulesep
    \endgroup
    \kern-\aboverulesep
  }%
}
\usepackage{mathtools}

\DeclarePairedDelimiter\norm{\lVert}{\rVert}

\title{Learning Query Expansion over the Nearest Neighbor Graph}

\addauthor{Benjamin Klein}{beni.klein@gmail.com}{1}
\addauthor{Lior Wolf}{wolf@cs.tau.ac.il}{1}

\addinstitution{
The Blavatnik School of Computer Science, \\ Tel Aviv University, Israel
}

\runninghead{Klein,Wolf}{Learning Query Expansion over the Nearest Neighbor Graph}

\begin{document}

\maketitle

\begin{abstract}
Query Expansion (QE) is a well established method for improving retrieval metrics in image search applications. When using QE, the search is conducted on a new query vector, constructed using an aggregation function over the query and images from the database. Recent works gave rise to QE techniques in which the aggregation function is learned, whereas previous techniques were based on hand-crafted aggregation functions, e.g., taking the mean of the query's nearest neighbors. However, most QE methods have focused on aggregation functions that work directly over the query and its immediate nearest neighbors. In this work, a hierarchical model, \oursfull (\ours), is presented, which is learned in a supervised manner and performs aggregation over an extended neighborhood of the query, thus increasing the information used from the database when computing the query expansion, and using the structure of the nearest neighbors graph. The technique achieves state-of-the-art results over known benchmarks.
\end{abstract}

\section{Introduction}

Most modern image search engines are based on the premise that an image can be effectively represented as a high dimensional feature vector (i.e. an embedding), 
such that the similarity between two images can be captured as the Euclidean distance between their corresponding embeddings. 
The embedding is usually obtained by a Convolutional Neural Network (CNN), trained to capture the semantic meaning of the image. A standard approach for finding similar images in a database of image embeddings, $D= \{ d_1, d_2, \dots, d_N \}$, for a given query embedding, $q$, is, therefore, to compute the Euclidean distance between the query embedding and the embedding of each image in the database, and rank the images in the database according to that distance. Given a fixed CNN used by an image search engine, further algorithmic improvements can be added to enhance the quality of the retrieval results, without changing the underlying CNN. A known group of such algorithmic improvements is called Query Expansion (QE). In QE, one enriches the query embedding, $q$, using the embeddings in the database, resulting in a new embedding for the query, $qe$. The new embedding, ${qe}$, is then compared to the embeddings in the database as before, resulting with a different ranking. It is important to note that when doing QE, the embeddings of the database images are not changed. A known and useful QE algorithm is Average Query Expansion~\cite{chum2007total}, in which the $K$ nearest neighbors of the query in the database, $\{ d_{j_{1}}, d_{j_{2}}, \dots, d_{j_{K}}\}$, are first found, and a simple average of their embeddings and the query is then computed, $\frac{1}{K+1}\cdot (q + \sum_{i=1}^{K}{d_{j_{i}}})$. The embedding is then normalized (e.g. using the $L_2$ norm), resulting in a new embedding for the query, $qe$. A further natural improvement to QE, is Database-Side Augmentation (DBA). In DBA, in addition to performing the expansion to the query embedding, the expansion is also performed (offline) to the the database images. Thus, as a pre-processing step, each image in the database is expanded using the same expansion technique, and the new embedding for each image in the database is stored instead of the original one. The image search is then performed between the expanded query embedding, and the expanded database embeddings.

QE algorithms can be divided into two groups. The first group are hand-crafted aggregation techniques~\cite{chum2007total, gordo2017learning, alphaqe}, in which the aggregation function that combines the information of the query and the database images is pre-defined. Such functions  usually have a few scalars as hyper-parameters. For example, the AQE can use different number of nearest neighbors $K$ when computing the mean. The second group are learned aggregation techniques~\cite{dqe, gordo2020attention},  which employ machine learning methods to define how to aggregate the information from the query embedding and the embeddings in the database. The recent Learnable Attention-based Query Expansion~\cite{gordo2020attention} (\theirs) uses a deep learning aggregation model trained on a dataset with a ranking loss, that receives the query embedding and the embeddings of its nearest neighbors in the database, and returns a new embedding for the query. 

The \oursfull (\ours) method, proposed here, extends the aggregation to be performed on an expanded neighborhood of the query, instead of limiting the aggregation to only its nearest neighbors. The method is a hierarchical one, where information is passed at $L$ stages. Each stage has a different learned aggregation function, and at each stage, a new embedding is computed for each image in  the expanded neighborhood of the query. The aggregator creates a new embedding for an image by aggregating the information from the embeddings of the node, and its nearest neighbors from the previous stage. Thus, A \ours model with two stages ($L=2$), aggregates the information hierarchically, such that the final embedding for the query, used for the QE, is composed from information aggregated from the nearest neighbors of the nearest neighbors of the query, as well as its immediate neighbors. A \ours model with $L$ stages, is aggregating the information hierarchically from $L$ neighbor hops from the query. Many known QE techniques can be seen as a special case of \ours with only one stage. As a natural further improvement, the \ours method can be applied to the database images as a DBA technique. Since the number of times that an aggregation function is applied for a single query grows exponentially with $L$, an improvement to the inference stage is suggested, reducing the computation cost to grow linearly in $L$, instead of exponentially. The proposed \ours method achieves state of the art results on several widely used retrieval benchmarks for both QE and DBA while having a good trade-off between the quality of the results and the time and memory resources required, and therefore, demonstrating the benefit of hierarchically aggregating information from extended neighbors of the query. 

\section{Related Work}

\paragraph{Query Expansion.} (QE) has long been a commonly used method in textual search engines~\cite{efthimiadis1996query}, 
in which a textual input query would be reformulated to improve the retrieval performance, by finding synonyms, stemming words, and other techniques. In the context of image search engines and information retrieval systems, which are based on deep learning representations, the QE method has a different meaning, in which the embedding of the query, $q$, and the embeddings of some images in the database are aggregated into a new embedding for the query, $qe$. It is then usually normalized, e.g., by applying $L_{2}$ normalization. 

Many successful and widely used QE algorithms~\cite{chum2007total, gordo2017learning, alphaqe} use hand-crafted aggregation methods that are defined by a few hyper-parameters. The Average Query Expansion (AQE)~\cite{chum2007total} computes a new embedding for the query, by using an aggregation that averages over the query embedding and its $K$ nearest neighbor image embeddings in the database, i.e. $\frac{1}{K+1}\cdot (q + \sum_{i=1}^{K}{d_{j_{i}}})$. 
The Average Query Expansion with Decay (AQEwD)~\cite{gordo2017learning} creates a new embedding for the query by an aggregation that computes a weighted sum over the query embedding and its $K$ nearest neighbor image embeddings in the database. 
The weights of the sum are a monotonically decreasing function of the nearest neighbors original ranking with respect to $q$, i.e. $q + \sum_{i=1}^{K} \frac{K - i}{K}\cdot d_{j_{i}}$. Therefore, AQEwD is giving more emphasis to the query and to the nearest neighbors that are ranked first. One can argue that both AQE and AQEwD do not scale well with the number of nearest neighbors, $K$. Since for AQE, as $K$ increases, the more similar the QE embedding is to a simple average over all the images in the database; While AQEwD has a decay factor such that the contribution of low ranked nearest neighbors converges to zero, it is still limited since the weight given to a sample depends only on its ranking, and is not a function of the query embedding and its neighbors. 
The Alpha Query Expansion ($\alpha QE$)~\cite{alphaqe} method has addressed this issue, by using an aggregation function that computes a weighted sum over the query embedding and its $K$ nearest neighbor image embeddings in the database, but differently from AQEwD which used weights that depend only on the ranking of the image, the $\alpha QE$ method is using weights which are a function of the cosine similarity between the query and its nearest neighbors, i.e. $q + \sum_{i=1}^{K} \left( \text{cosine-sim} \left( q, d_{j_{i}} \right) \right)^{\alpha} {d_{j_{i}}}$, where $\alpha$ is an additional hyper-parameter. Thus, with  the $\alpha QE$ method, a query that has many nearest neighbors that have a small distance to it, will have a QE that depends on more nearest neighbors, than another query for which most of its nearest neighbors are farther away. A previous work~\cite{iscen2017efficient} has suggested a QE method based on diffusion~\cite{donoser2013diffusion} in which information is propagated on the nearest neighbors graph. The propagation procedure itself does not utilize supervised learning, and while powerful, the method does not scale well and can require significant amount of resources for large graphs~\cite{iscen2018fast}. More efficient methods based on diffusion and spectral methods were suggested~\cite{iscen2018hybrid,iscen2018fast} but with the cost of a slight degradation in performance or an increase in memory resources. 
Explore-Exploit Graph Traversal (EGT)~\cite{chang2019explore} is using a powerful and efficient re-ranking technique that does not involve learning and also utilizes the nearest neighbor graph. EGT traverses the graph, starting from the query, while making a trade-off between taking images which are nearby the query (exploit) and between extending the search farther from the query (explore). 
A recent work~\cite{liu2019guided} has proposed a method for learning how to propagate information on the nearest neighbor graph. Similarly to \ours, this method is employing a graph neural network~\cite{kipf2016semi} to propagate information on the graph. Differently than \ours, this method is using an unsupervised loss and does not utilize supervision. Additionally, its inference has a dependency that grows as a function of $O(K^L)$, where $K$ is the number of nearest neighbors and $L$ is number of nearest neighbor hops used in the aggregation. 
In contrast, as discussed in  Subsection~\ref{subsec:efficientinference}, \ours can utilize efficient inference which requires only $L$ calls to the aggregator, where each call to the aggregator utilizes information from $K$ items in the database. The \ours can complement most of these methods (e.g., EGT~\cite{chang2019explore}), by first computing new embeddings using \ours and then applying these methods. The other group of QE methods employ Machine Learning algorithms to define the aggregation function. The Discriminative Query Expansion (DQE)~\cite{dqe}, takes the high ranked images of a given query as positive data points, and low ranked images as negative data points, and trains a linear SVM. The distance of a sample from the linear separator is then used by the aggregation function. 

A recent QE method, \theirs~\cite{gordo2020attention}, has defined an aggregation function that is a fully differentiable deep learning model, which receives the embeddings of the query and its nearest neighbors and returns the expanded query. Following the success of attention models and of transformers~\cite{gordo2020attention}, the chosen deep learning model is the encoder part of a transformer. 
\vspace{-12pt}
\paragraph{Graph Neural Networks.}

For quite a while, deep learning methods have been achieving state of the art results on visual and textual tasks, but it is only recently that deep learning methods have been successfully employed and achieved state of the art results on structured data, such as graphs. Many of the first Graph Neural Network (GNN) techniques have focused on learning node embeddings~\cite{perozzi2014deepwalk, grover2016node2vec} given a structure of a graph. DeepWalk~\cite{perozzi2014deepwalk} has used random walks over the graph to define a sequence of nodes, which can then be used to train a Skip-gram model~\cite{mikolov2013efficient}, resulting with an embedding for each node in the graph. While these methods can efficiently learn embeddings for the nodes of a given graph, they are not inductive, in the sense that the parameters learned by the model, are the node embeddings themselves, and those are not transferable to another graph or dataset. The parameters of the \ours model proposed in this paper are trained on one dataset, and are then evaluated on other datasets. Therefore, it is necessary that the hierarchical aggregation approach used by the model to be inductive. A few methods~\cite{hamilton2017representation, ying2018graph, hamilton2017inductive} have been suggested that are inductive, and can be learned on one graph and applied later to other graphs. In those methods, one usually starts from an initial embedding for each node in the graph, and learns a local operator that aggregates information from a local area around the node, resulting in an operator that is transferable to other graphs. While many of these graphs were employed on natural graphs, such as citation networks~\cite{sen2008collective}, the \ours method described in this paper is learning a hierarchical model on a graph defined by the $K$ nearest neighbors of each item in a database of images. The hierarchical model learns how to aggregate information from $L$ hops of nearest neighbors with respect to the query, and is fully transferable from a nearest neighbor graph defined on one dataset, to another unseen nearest neighbor graph defined on a dataset unseen at training time.

\renewcommand{\AlCapSty}[1]{\normalfont\footnotesize {\textbf{#1}}\unskip} 
\begin{figure}[t!]

\begin{minipage}[t]{0.42\textwidth}
\begin{algorithm}[H]
\label{algo:agg}
\footnotesize
\SetAlgoLined
\DontPrintSemicolon
 \KwIn{$\left(v, d_{j_{1}}, \dots, d_{j_{K}} \right)$}
 \KwOut{A new embedding for $v$, $v^{*}$}
 $v\gets v + \text{positional-embedding}(0)$\;
  \For{$i\gets1$ \KwTo $K$}{
    $d_{j_{i}}\gets d_{j_{i}} + \text{positional-embedding}(i)$\;
 }
 $\small( \tilde{v}, \tilde{d_{j_{1}}}, \dots, \tilde{d_{j_{K}}} \small) \gets \text{encoder}\left(v, d_{j_{1}}, \dots, d_{j_{K}} \right)$\;
  $sim_{0} \gets 1$\;
   \For{$i\gets1$ \KwTo $K$}{
        $sim_{i} \gets \text{cosine-sim}(\tilde{v}, \tilde{d_{j_{i}}})$\;
   }
   
   $v^{*} \gets sim_{0} \cdot v + \sum_{i=1}^{K} sim_{i} \cdot d_{j_{i}}$\;
   $v^{*} \gets \frac{v^{*}}{\norm{v^{*}}}$\;
    \Return $v^{*}$

 \caption{Aggregation Function}
\end{algorithm}
\end{minipage}
\hfil
\begin{minipage}[t]{0.50\textwidth}

\begin{algorithm}[H]
\footnotesize

\label{algo:GEQ}
\SetAlgoLined
\DontPrintSemicolon
 \KwIn{Query image, $q$; a function, $NN_{K}(v)$; }
 \myinput{Database Images, $DB= \{ d_1, d_2, \dots, d_N \}$}
 \KwOut{$qe$, The QE Embedding of $q$, }
  $S^{L} \gets \{ q \} $\; 
  \For{$ i\gets 1$ \KwTo $L$}{
    \ForEach{$u \in S^{L-i+1}$}{
    $S^{L-i} \gets S^{L-i+1} \cup NN_{K}(u)$
    }
  }
  
  \ForEach{$u \in S^{0}$}{
    $u^{0} \gets u$
  }
  
   \For{$ i\gets 1$ \KwTo $L$}{
    \ForEach{$u \in S^{i}$}{
    $\left (d_{j_{1}}, \dots, d_{j_{K}} \right) \gets NN_{K}(u)$\;
    $u^{i} \gets agg_{i} \left(u^{i-1}, d_{j_{1}}^{i-1}, \dots, d_{j_{K}}^{i-1} \right)$
    }
  }
  
  $qe \gets q^{L}$\;
 \Return $qe$
 \caption{Hierarchical Query Expansion}
\end{algorithm}
\end{minipage}
\end{figure}

\section{Graph Query Expansion}
Let $\phi$ be an image feature extractor that transforms an input image, $x$, into a semantic embedding, $\phi(x) \in \mathbb{R}^F$ (in practice $\phi$ is a pre-trained CNN). In the following sections, any reference to an image will refer to its embedding. Given a database of images, $D = \{d_1, d_2, \dots d_N\}$, and an image, $v$, we define $NN_{K}(v)$ to be the $K$ nearest neighbors in the database with respect to $v$ according to the Euclidean distance in the embedding space. The computation is performed using an hierarchical aggregator, defined by a hyper-parameter, $L$, that defines the number of nearest neighbor hops taken from the query. The hierarchical computation is performed on a local directed graph, $G$, that is constructed as follows; First the query image $q$ is added to the graph. Then, for each $i= \{ 1, 2, \dots, L \}$, a directed edge is added from each node $v$, that is already in the graph, to each of its $K$ nearest neighbors, defined by $NN_{K}(v) = \left( d_{j_{1}}, \dots, d_{j_{K}} \right)$. For example, when $L=1$, only a single hop is considered, and the new embedding for the query is a function of its only $K$ nearest neighbors. In another scenario, when $L=2$, two hops are considered, and the new embedding for the query depends on the nearest neighbors of the query as well as the nearest neighbors of the nearest neighbors of the query, making it a function of at most $O(K^2)$ different database images. The hierarchical aggregation computation is fully described in Algo~\ref{algo:GEQ}. The computation is done at $L$ steps, where at each step a different aggregator, $agg_i$, is applied to a node and its nearest neighbors in the graph, and returns a new embedding for the node. Each aggregator is a transformer-encoder, similar to the one used by \theirs~\cite{gordo2020attention}. For completeness, the aggregator function is described in Algo~\ref{algo:agg}. The computation starts by defining $L+1$ sets  (lines $1-6$ in Algo~\ref{algo:GEQ}), where the $S^{L}$ set contains only the query, and the $S^{L-i}$ set contains the query, and all the database images that are within at most $i$ nearest neighbor hops from it. These sets define which images will be aggregated by the $i-th$ aggregator at each step of the computation. Thus, only the images in $S^{L-i}$, which are all the images with distance of at most $i$ neighbor hops from the query will be aggregated by the $i-th$ aggregator. In lines $7-9$ in Algo~\ref{algo:GEQ}, the initial embedding, $u^{0}$, of each image, $u$, in $S^0$ (that contains all the images within $L$ neighbor hops from the query) is set to $u$, i.e. the initial embedding of every image is set to the embedding of the image resulting from the feature extractor, $\phi(x)$. The hierarchical aggregator then recursively computes new embeddings for the images (lines $10-15$ in Algo~\ref{algo:GEQ}). At $i-th$ step of the recursion, the $i-th$ representation, $v^{i}$ of each image, $v$, in $S_i$ is set to: ${v}^{i} = agg_i \left(v^{i-1}, d_{j_{1}}^{i-1}, \dots, d_{j_{K}}^{i-1} \right)$ by applying the $i-th$ aggregator, $agg_{i}$, where $v^{i-1}$ is the embedding of image, $v$, from the previous step of the computation, and $\left( d_{j_{1}}^{i-1}, \dots, d_{j_{K}}^{i-1} \right)$ are the embeddings from the previous step of the computation for all the $K$ nearest neighbors of $v$, $\left( d_{j_{1}}, \dots, d_{j_{K}} \right)$. The QE representation of the query, $q$ is equal to $q^{L}$, the representation of $q$ at the $L$ step of the recursion and it is returned in line $17$ in Algo~\ref{algo:GEQ}.

\vspace{-12pt}
\paragraph{Training}
\label{para:training}
In all the experiments, the state of the art CNN provided by \cite{alphaqe} is the feature extractor, $\phi(x)$, used to extract embeddings from the images. The CNN is based on a Resnet-101 architecture~\cite{he2016deep} (not including the last layer), followed by generalized-mean pooling and a whitening layer. The CNN is trained on the Google Landmarks 2018 Dataset~\cite{noh2017large} and its output is a $2048$ dimensional vector. Each image is passed through the CNN at three scales $\left(1, \sqrt{2}, 1/\sqrt{2} \right)$ which are then averaged, followed by $L_{2}$-normalization. The training is performed on the training data of rSfM120k. Tuples $(q, p, n)$, are constructed from the training set images and labels, where $q$ is a query image, $p$ is a positive image which is semantically related to $q$, and $n$ is a negative image which is not semantically related to $q$. The query, $q$, is passed to the hierarchical aggregator, Algo~\ref{algo:GEQ}, which constructs a dynamic computational graph, where the leaves are the $0$-th level of the computational graph, and contain the original embeddings of images that are $L$ hops away from the query. Then, for $i=1, \ldots, L$, each internal node in the $i$-th level of computational graph is the result of applying $agg_i$ (a differentiable Neural Network) on the relevant $K+1$ nodes in level $i-1$. The overall network consists of $O(K^{L})$ many nodes, and is trained end-to-end (with all the aggregators applied at the $i$-th step sharing parameters). 
Finally, the expanded query, $qe$, is passed together with $p$, and $n$, to a Contrastive Loss~\cite{hadsell2006contrastive}, and the parameters of the hierarchical aggregator model are then updated by applying back-propagation~\cite{rumelhart1986learning}. The training procedure derives from the PyTorch implementation\footnote{https://github.com/filipradenovic/cnnimageretrieval-pytorch} of \cite{alphaqe}. For each pair of a query and a positive sample, five negative samples are selected from a pool of $20000$ images which is updated every $2000$ training iterations. The hard mining of negative samples is done with respect to the expanded query. Since the CNN used to extract the embedding of each image is already very powerful, many of the negative pairs have $0$ contribution to the Contrastive Loss, which emphasize the importance of hard mining them. Each aggregator at each step of the hierarchical computation is a transformer-encoder, as in \theirs. Specifically, each aggregator is a transformer-encoder with $64$ heads, three layers, and feed forward dimensionality of $2048$. A positional embedding is added to each aggregator, as described in Algo~\ref{algo:agg}. The model is trained for $40$ epochs, with a batch size of $64$, using the Adam optimizer with a learning rate of $5e{-5}$, and weight decay of $1.5e{-6}$. The margin used for the Contrastive Loss function is $0.71$. The hyper-parameters are selected on the validation data of rSfM120k, and the best model with respect to the $mAP$ on the validation data of rSfM120k is then used for the evaluation of the tests sets: \roxf, \rpar, and the \rmil distractors. The chosen \ours model has $L=2$ steps, and $K=44$ nearest neighbors are used by each aggregator. Therefore, the upper bound on the number of database images that participate in the computation of the expanded query, $qe$, is $K^2 + K = 1980$. The upper bound on the number of times the transformer-encoder is applied for a single query, $O(K^{L-1})$, grows exponentially with $L$, and therefore, both the time and memory resources required to train the model become a bottleneck for large values of $L$ and $K$. This limits the experiments from learning \ours for $L=3$ for a sufficiently large value of $K$. By using a GPU with a larger memory, an experiment with $L=3$ and $K=36$ was conducted in which similar results to $L=2$ were obtained but without surpassing them ($K > 36$ was not tested due to reaching the GPU memory limit).
For efficient computation, the $K$ nearest neighbors of each query image and each database image are pre-computed and cached.
\definecolor{Gray}{gray}{0.9}

\begin{table}[t]
\label{tab:allresults}
\begin{center}
\footnotesize
\def\cw{1cm}
\newcolumntype{L}[1]{>{\raggedright\let\newline\\\arraybackslash\hspace{0pt}}m{#1}}
\newcolumntype{C}[1]{>{\centering\let\newline\\\arraybackslash\hspace{0pt}}m{#1}}
\newcolumntype{R}[1]{>{\raggedleft\let\newline\\\arraybackslash\hspace{0pt}}m{#1}}
\begin{tabular}{L{2.6cm}C{0.5cm}C{0.5cm}C{0.5cm}C{0.5cm}C{\cw}C{0.5cm}C{0.5cm}C{0.5cm}C{0.65cm}}
\toprule
 & \multicolumn{2}{c}{\rox} & \multicolumn{2}{c}{\rox~+~\rmil} & \multicolumn{2}{c}{\rpa} & \multicolumn{2}{c}{\rpa~+~\rmil} & \\
\cmidrule(lr){2-3} \cmidrule(lr){4-5} \cmidrule(lr){6-7} \cmidrule(lr){8-9}
  & M & H & M & H & M & H & M & H & Mean \\
\toprule
 \multicolumn{10}{c}{\cellcolor{Gray}{}QE} \\
 \midrule
No QE & 67.3 & 44.3 & 49.5 & 25.7 & 80.6 & 61.5 & 57.3 & 29.8 &  52.0\\
\midrule
 AQE~\cite{chum2007total} & 72.3 & 49.0 & 57.3 & 30.5 & 82.7 & 65.1 & 62.3 & 36.5 & 56.9 \\
 \midrule

AQEwD~\cite{gordo2017learning} & 72.0 & 48.7 & 56.9 & 30.0 & 83.3 & 65.9 & 63.0 & 37.1 & 57.1 \\
\midrule
DQE~\cite{dqe}  & 72.7 &  48.8 & 54.5 & 26.3 & 83.7 & 66.5 & 64.2 & 38.0 &  56.8 \\
\midrule
$\alpha\text{QE}$~\cite{alphaqe} & 69.3 & 44.5 & 52.5 & 26.1  & 86.9 & 71.7  & 66.5 & 41.6 &  57.4 \\
\midrule
EGT~\cite{chang2019explore} & 66.1 & 44.5 & - & -  & 82.5 & 68.4  & - & - &  - \\
\midrule
\theirs~\cite{gordo2020attention} & 73.4 &  49.6 & 58.3 & 31.0 & 86.3 & 70.6 & 67.3 & 42.4 &  59.8 \\
\midrule
\ours \textbf{(ours)} & \textbf{74.1} &  \textbf{51.0} & \textbf{59.4} & \textbf{32.7} & \textbf{87.4} & \textbf{72.4} & \textbf{69.5} & \textbf{45.3} &  \textbf{61.4} \\

\toprule
\multicolumn{10}{c}{\cellcolor{Gray}{}DBA} \\
 \midrule
 DBA + AQE~\cite{chum2007total} & 71.9 & 53.6 & 55.3 & 32.8 & 83.9 & 68.0 & 65.0 & 39.6 & 58.8 \\
 \midrule

DBA + AQEwD~\cite{gordo2017learning} & 73.2 & 53.2 & 57.9 & 34.0 & 84.3 & 68.7 & 65.6 & 40.8 & 59.7 \\
\midrule
DBA + DQE~\cite{dqe}   & 72.0 & 50.7 & 56.9 & 32.9 & 83.2 & 66.7 & 65.4 & 39.1 &  58.4 \\
\midrule
DBA + $\alpha\text{QE}$~\cite{alphaqe} & 71.7 & 50.7 & 56.0 & 31.5 & 87.5 & 73.5 & 70.6 & 48.5 &  61.3 \\
\midrule
DBA + \theirs~\cite{gordo2020attention} & 74.0 & 54.1 & 60.0 & 36.3 & 87.8 & 74.1 & 70.5 & 48.3  & 63.1 \\
\midrule
DBA + \ours \textbf{(ours)} & \textbf{75.3} & \textbf{56.1} & \textbf{60.3} & \textbf{36.4} & \textbf{88.6} & \textbf{75.2} & \textbf{72.9} & \textbf{51.6}  & \textbf{64.5}\\

\bottomrule
\end{tabular}
\end{center}
\vspace{-12pt}
\caption{
Performance evaluation of the mean average precision (mAP) on the QE task (top section) and DBA task (bottom section). Results are reported on \roxf (\rox) and \rpar (\rpa) with and without 1 million distractors (\rmil). Our method, \ours, \theirs and EGT employ the validation part of rSfM120k for hyper-parameter selection. All other methods optimized their hyper-parameters directly on the mAP over all $M$ and $H$ queries of the test sets of \roxf and \rpar, which provides them with an advantage. Results for EGT are not provided for \rox~+~\rmil and \rpa~+~\rmil due to high memory requirements of the code provided for EGT. \vspace{-11pt}
}
\label{tab:baseline_comparison_qe}
\end{table}

\vspace{-8pt}
\paragraph{Database-Side Augmentation} (DBA)~\cite{dqe} A further intuitive improvement in the retrieval performance, can be achieved by applying the model on the database images as well, by running Algo~\ref{algo:GEQ} on each database image. The process is done offline, by computing the expanded version of each image in the database, using the \ours model, and storing the new expanded embedding, instead of the original one. 
\vspace{-12pt}
\paragraph{Efficient Inference}
\label{subsec:efficientinference}
At first glance, it may seem that using \ours dramatically increases the computational resources for computing the QE for a query, $q$, since the computation requires running transformer-encoder $O(K^{L-1})$ times, which grow exponentially with $L$, in contrast to \theirs~\cite{gordo2020attention} which requires applying a single transformer-encoder. Upon further review of the hierarchical expansion (Algo~\ref{algo:GEQ}), once trained, one can apply \ours to any query, $q$, using only $L$ calls to a transformer-encoder aggregator, by storing additional information for each image in the database. The main observation is that for every image in the database, $v$, the value of $v^{i}, i=1 \dots L$ depends only on the images in the database, and does not depend on query. Therefore, one can pre-process, $v^{i}, i=1 \dots L-1$, for every image, $v$, in the database, and store them in addition to the image embedding. Consequentially, increasing the memory required for storing the database by a factor of $O(L)$.

\section{Experiments}
This section describes in detail how the model is trained and evaluated. For a fair evaluation, the experiments follow the protocols of \theirs~\cite{gordo2020attention}.

\vspace{-15pt}
\paragraph{Datasets} The rSfM120k dataset~\cite{alphaqe} contains the training data used for training \ours, and the validation data on which the hyper-parameters of \ours are tuned. The training data includes $91642$ images, each belonging to one of $551$ classes. The class information is used for selecting positive and negatives samples for a given query, as described in Section~\ref{para:training}. The validation data includes $6403$ images, each belonging to one of $162$ classes which are disjoint from the classes in the training data. The $1691$ query images which are a subset of the $6403$ images, are used for computing the mean average precision~\cite{philbin2007object} (mAP) which is the chosen validation metric. The Revisited Oxford (\roxf) and the Revisited Paris (\rpar) datasets~\cite{radenovic2018revisiting} are the revisited versions of the corresponding well-known datasets of Oxford~\cite{philbin2007object} and Paris~\cite{philbin2008lost}. The \roxf dataset contains $4993$ database images and $70$ query images, and the \rpar dataset contains $6322$ database images and $70$ query images. Each query image in both datasets is labeled as either Easy (E), Medium (M) or Hard (H). Since the results on the Easy query images are known to be saturated, it is a common practice to report metrics only on the Medium and Hard query images. The \rpar and \roxf datasets are both disjoint from rSfM120k. Since \rpar and \roxf are relatively small datasets, they do not reflect large scale image search applications, where the size of the database is much larger. A dataset of 1 millions distractors~\cite{radenovic2018revisiting} (\rmil) can be added to the database images of \rpar and \roxf, increasing the difficulty of the retrieval task. The distractor images do not match any of the query images in \rpar and \roxf. 

\begin{figure}
\floatbox[{\capbeside\thisfloatsetup{capbesideposition={left,top},capbesidewidth=3.6cm}}]{figure}[\FBwidth]
{\caption{\scriptsize	
The mean over the mAP of the $M$ and $H$ queries of the \roxf and \rpar datasets in the QE task, as a function of the number of neighbors used by each method. The results for \ours method are obtained for $L=2$. Notice that in contrast to other methods that are using $K$ database images in the QE computation of a single query, the methods \ours, AQE-G and $\alpha\text{QE}$-G are using $O(K^2)$ different database images. Since the hyper-parameters of \ours are chosen on the validation data of rSfM120k, the chosen value of $K=44$ is not the $K$ that maximizes the mAP value.}\label{fig:knn}}
{\includegraphics[width=9cm]{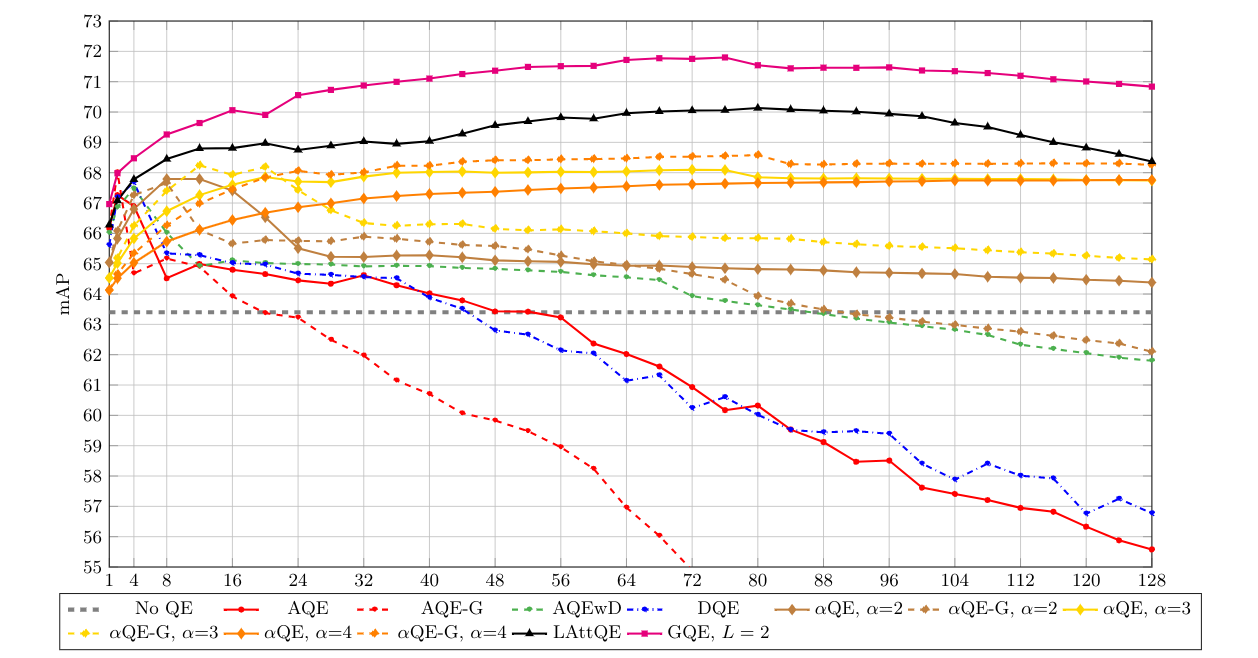}}
\end{figure}

\subsection{Results}
\label{sec:results}

The performance of the \ours method is compared to \theirs~\cite{gordo2020attention},
Average Query Expansion (AQE) ~\cite{chum2007total}, Average Query Expansion with Weight Decay (AQEwD)~\cite{gordo2017learning}, Discriminative QE (DQE)~\cite{dqe}, and $\alpha\text{-weighted}$ QE~\cite{alphaqe}. The results reported for these methods are taken as is from~\cite{gordo2020attention}. Another comparison is made for EGT~\cite{chang2019explore} by using the source code provided. Notice that for a fair comparison, the EGT method is applied to the same embeddings used by all the other methods in the experiment, and therefore the results reported here for EGT are different than those reported in~
\cite{chang2019explore} where the images were represented by different features. While the parameters and hyper-parameters of \theirs~\cite{gordo2017learning}, EGT~\cite{chang2019explore}, and \ours are selected on the validation data of rSfM120k, the hyper-parameters of all the other methods are selected directly on the mean performance of the test sets, giving them a slight advantage.
\vspace{-18pt}

\paragraph{QE.} In the QE protocol, where the expansion is applied only to the queries and not to the database, \ours achieves state-of-the-art results, and improves the mAP for both the \roxf and \rpar datasets both when not using the \rmil distractors and when adding them, as shown in Table~\ref{tab:baseline_comparison_qe}. The mean mAP over both the Medium and Hard queries of the \roxf and \rpar datasets as a function of the number of the nearest neighbors, $K$, when compared to other methods is shown in Fig~\ref{fig:knn}. To further understand the contribution of the hierarchical aggregation, additional experiments in which the hierarchical aggregation is paired with other QE techniques are presented. Specifically, AQE-G and $\alpha\text{QE}$-G in Fig~\ref{fig:knn} are obtained by combining hierarchical aggregation with AQE and $\alpha\text{QE}$. It is worth noting, that while $K$ different database images are participating in the QE of all the other methods, in \ours, AQE-G, and $\alpha\text{QE}$-G at most $K^2+K$ different database images can participate in the computation of a single QE. Notice that since \ours hyper-parameters were chosen on the validation data of rSfM120k, the value of $K=44$ is not the $K$ that maximizes the mAP value in Fig~\ref{fig:knn}. As seen, combining the hierarchical aggregation with other QE techniques (e.g., AQE) is inferior to learning the aggregation explicitly as done in the proposed \ours approach. 
\begin{figure}[t]
\includegraphics[width=12cm]{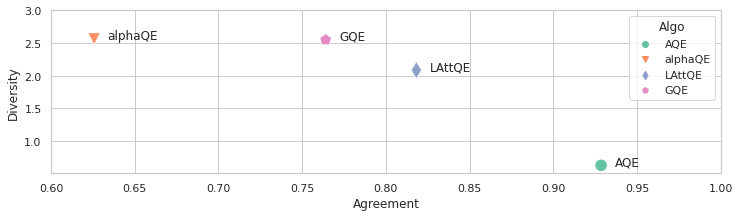}
 \vspace{-12pt}
\caption{Comparison of the proposed metrics, \emph{Agreement} and \emph{Diversity}, on the \roxf dataset across different QE methods. \vspace{-16pt}
}

\label{fig:metrics}
\end{figure}

For further evaluating the contribution and benefit of using the information from an extended neighborhood of the query, the chosen \ours model with $L=2$ is modified such that first aggregator simply returns the node, and ignores its neighbors, i.e., $agg_{1}\left(v, d_{j_{1}}, \dots, d_{j_{K}} \right) = v$. Therefore, the resulting model from this modification, collapsed \ours, is computationally equivalent to \theirs. The results obtained are presented in Table~\ref{tab:gqemodification}, and are very similar to the results obtained by \theirs, which further supports our claim that using the information from an extended neighborhood of the query, beyond its immediate nearest neighbors does contribute to the improvement in the retrieval performance.

The query expansion for most methods and specifically for AQE, $\alpha$QE, \theirs, and \ours can be seen as a weighted average of embeddings from the database, $qe = \sum_{i} w_i \cdot v_i $ where $w_i \geq 0 $. Notice that in the case of AQE, $\alpha$QE, and \theirs the sum is defined over the nearest neighbors of the query, and in the case of \ours the sum is defined over all the nearest neighbors within two hops from the query. Additionally, the weights in \theirs and \ours are the result of applying a non linear function on the query and its neighbors (i.e. the neural network). We introduce two metrics for QE that provide further insights to the difference between the QE methods. The first metric, called \emph{Agreement}, is defined as the ratio between the sum of the weights of images that share the same label as the query and  the total sum of weights, i.e., $Agreement(qe) = \left( \sum_{y_q = y_i} w_i \right) / \left( \sum w_i \right)$, where $y_q$ is the label of the query, and $y_i$ is the label of the $i$-th image. The second metric, called \emph{Diversity}, is defined as the entropy of the probability distribution over the weights that belong to the same label as the query (the probability distribution is obtained by normalizing the weights) . Our hypothesis is that a powerful QE algorithm needs to have both a high \emph{Agreement} value (i.e., for many of the items in the sum to have the same label as the query) and a high \emph{Diversity} value (i.e., that the weight is not concentrated on a small number of samples). We analyze the results on the \roxf dataset from the perspective of these two metrics in Figure~\ref{fig:metrics}. The parameters of each algorithm in the analysis are the same ones for which the results in Table~\ref{tab:baseline_comparison_qe} were obtained. As shown, while AQE has the highest \emph{Agreement} value it also has the smallest value of \emph{Diversity} which can explain its low performance. While $\alpha$QE has a high \emph{Diversity} value it also has the smallest value of \emph{Agreement}. The method of \theirs also suffers from a small value of \emph{Diversity}. Our conclusion is that \ours provides a good balance between having a high value of \emph{Agreement} and a high value of \emph{Diversity} which enables the method to surpass the others methods.
\vspace{-8pt}
\paragraph{DBA.} In DBA, one applies the expansion model to both the query and the dataset with the hope of obtaining further improvement in the retrieval performance, when compared to preforming the expansion only to the query. Similar to the findings of~\cite{gordo2020attention}, applying \ours directly to the database did not result in a significant improvement. The solution proposed in~\cite{gordo2020attention} is for the aggregation function to apply a tempered softmax on the similarity weights, i.e., dividing the $sim$ vector described in Algorithm~\ref{algo:agg} by a hyper-parameter, $T > 0$, and applying softmax. Thus, making the $sim$ vector either more uniform by using large values of $T$, which in the extreme case is equivalent of doing AQE, or making it closer to a one hot encoding by using small values of $T$, which in the extreme case is equivalent to not doing any expansion and returning $v$. Since \ours is composed of multiple aggregators, a different value $T_{i}$ is used for each aggregator. Therefore, the \ours model used in our experiments has a hyper-parameter $T_{1}$ which is associated with the aggregator of the first level, and a hyper-parameter $T_{2}$ which is associated with the aggregator of the second level. Similarly to the other methods, a different value of $K$ is used for the database expansion, $K_{DBA}$, than the one used for the query. Following \theirs, the hyper-parameters $T_{1}$, $T_{2}$, and $K_{DBA}$ are selected by freezing the \ours model chosen for the QE task, and optimizing $T_{1}$, $T_{2}$, and $K_{DBA}$. With those modifications, our \ours method achieves state of the art results for DBA as well, for both the \roxf and \rpar datasets, with and without adding the \rmil distractors as shown in Table~\ref{tab:baseline_comparison_qe}.

\begin{table}[t]
\begin{center}
\footnotesize
\def\cw{0.5cm}
\newcolumntype{L}[1]{>{\raggedright\let\newline\\\arraybackslash\hspace{0pt}}m{#1}}
\newcolumntype{C}[1]{>{\centering\let\newline\\\arraybackslash\hspace{0pt}}m{#1}}
\newcolumntype{R}[1]{>{\raggedleft\let\newline\\\arraybackslash\hspace{0pt}}m{#1}}
\begin{tabular}{lcccc}
\toprule
 & \multicolumn{2}{c}{\rox} & \multicolumn{2}{c}{\rpa} \\
\cmidrule(lr){2-3} \cmidrule(lr){4-5}
  & M & H  & M & H \\

\midrule
\theirs~\cite{gordo2020attention} & 73.4 & 49.6 & 86.3 & 70.6 \\
Collapsed \ours & 72.0 & 49.3 & 86.3 & 70.5  \\
\ours (ours) & \textbf{74.1} & \textbf{51.0} & \textbf{87.4} & \textbf{72.4}  \\
\bottomrule

\end{tabular}
\vspace{-10pt}

\end{center}
\caption{
A Collapsed \ours is created by modifying \ours with $L=2$ after it was trained; The aggregator at the first level of the model is changed to simply return the input node, ignoring the $K$ nearest neighbors. Thus, the Collapsed \ours is functionally equivalent to \theirs. As shown, the results for \theirs and the Collapsed \ours are very similar. This further supports the claim that information from expanded neighborhood of the query contribute to the QE and to the retrieval performance. The mAP results are presented for the QE task for the \rpar and \roxf datasets.  \vspace{-15pt}
}
\label{tab:gqemodification}
\end{table}

\section{Conclusions} \label{sec:conclusion}
The proposed \ours method has demonstrated the benefits of doing QE on an extended neighborhood of the query, instead of limiting the QE to the immediate nearest neighbors of the query. By formulating the aggregation procedure over the Nearest Neighbors Graph, as a Graph Neural Network model, and using state of the art aggregation models~\cite{gordo2020attention}, the technique achieves state of the art results in both the QE and DBA tasks. 
Since, the memory resources required for training \ours are increasing exponentially with $L$ as discussed in Section~\ref{para:training}, our experiments were limited to either using $L=2$ or to using $L=3$ with a low value of $K$. Further improvements may be achieved by using aggregation models which require less memory~\cite{choromanski2020rethinking, tay2020long}, thus enabling experimenting with larger values of $L$, and using information from farther neighbors. 

\section{Acknowledgments}
This project has received funding from the European Research Council (ERC) under the European Unions Horizon 2020 research and innovation programme (grant ERC CoG 725974). The contribution of the first author is part of a Ph.D. thesis research conducted at Tel Aviv University.

\bibliography{egbib}
\end{document}